\documentclass[journal, twoside]{IEEEtran}
\usepackage[english]{babel}
\usepackage{amsmath}
\usepackage{amssymb}
\usepackage{amsfonts}
\usepackage{amsthm}
\usepackage{algorithm}
\usepackage{algpseudocode}
\usepackage{graphicx}
\usepackage{dblfloatfix}
\usepackage{subcaption}
\usepackage{epsfig}
\usepackage{tensor}
\usepackage{comment}
\usepackage{color}

\usepackage[dvipsnames]{xcolor}
\definecolor{rephrase}{RGB}{219, 10, 10}

\usepackage{tablefootnote}
\usepackage{diagbox}
\usepackage{gensymb}
\usepackage{environ}         
\usepackage{etoolbox}        
\usepackage[flushleft]{threeparttable} 
\graphicspath{{figures/}}
\DeclareGraphicsExtensions{.eps}
\theoremstyle{definition}

\DeclareMathOperator*{\argmin}{arg\,min}

\usepackage{tablefootnote}

\begin{document}

\title{\LARGE \bf A Neurorobotic Embodiment for Exploring the Dynamical Interactions of a Spiking Cerebellar Model and a Robot Arm During Vision-based Manipulation Tasks}

\author{Omar Zahra, David Navarro-Alarcon and Silvia Tolu%
\thanks{O. Zahra and D. Navarro-Alarcon are with The Hong Kong Polytechnic University, Department of Mechanical Engineering, Kowloon, Hong Kong. Corresponding author e-mail: {\texttt{\small dna@ieee.org}}.}
\thanks{S. Tolu is with Technical University of Denmark, Department of Electrical Engineering, Copenhagen, Denmark.%
}}

\bstctlcite{IEEEexample:BSTcontrol}

\maketitle
\thispagestyle{empty}
\pagestyle{empty}

\begin{abstract}
 While the original goal for developing robots is replacing humans in dangerous and tedious tasks, the final target shall be completely mimicking the human cognitive and motor behaviour. Hence, building detailed computational models for the human brain is one of the reasonable ways to attain this. The cerebellum is one of the key players in our neural system to guarantee dexterous manipulation and coordinated movements as concluded from lesions in that region. Studies suggest that it acts as a forward model providing anticipatory corrections for the sensory signals based on observed discrepancies from the reference values. While most studies consider providing the teaching signal as error in joint-space, few studies consider the error in task-space and even fewer consider the spiking nature of the cerebellum on the cellular-level. In this study, a detailed cellular-level forward cerebellar model is developed, including modeling of Golgi and Basket cells which are usually neglected in previous studies. To preserve the biological features of the cerebellum in the developed model, a hyperparameter optimization method tunes the network accordingly. The efficiency and biological plausibility of the proposed cerebellar-based controller is then demonstrated under different robotic manipulation tasks reproducing motor behaviour observed in human reaching experiments.

\end{abstract}

\section{INTRODUCTION}
Neurorobotics is getting more attention nowadays not only for improving the performance of robot controllers but also to reveal some of the mysteries about how our brains work \cite{chiel1997brain,rucci2007integrating}. Limitation of state-of-the-art techniques to monitor spiking activity of all neurons in the brain makes it necessary to develop accurate computational models to verify theories about neural brain mechanisms \cite{krichmar2002machine}. Also, the mutual benefit that derives from the joint research in neuroscience and robotics fields \cite{edelman2007learning} enables the development of adaptive biologically inspired controllers that can be used as a basis to explain mechanisms of learning and enhance the performance of robots. Hence, the motor system lies among the most studied for the common interest in both fields. Each of the brain regions contributing to the motor control have distinctive features leading to different roles in the control process.

This study is concerned with modeling an essential complex region in our motor system, the cerebellum \cite{eccles2013cerebellum,buckner2013cerebellum}. The cerebellum is well known to help achieve fine motor control and precise timing and coordination of the movement of the joints to achieve dexterous motion \cite{porrill2013adaptive}. That was proven by studies of patients with lesions in the cerebellum \cite{kandel2000principles} suffering from clumsy staggering movements similar to a drunken behaviour. In robotics field, incorporating a cerebellar model contributes to enhancing the accuracy and precision of robot movements which is critical in many robotic applications like surgery \cite{luque2011adaptive,tolu2013adaptive}.\\
Various models \cite{wolpert1998internal} were built for the cerebellum based upon different theories for its role in our movements. One theory is that cerebellum acts as an inverse model that provides corrections for the motor commands \cite{kawato1987hierarchical}. Another theory is that it acts instead as a forward model to improve the sensory predictions \cite{ishikawa2016cerebro}. Further theories were introduced as well to combine the merits of both forward and inverse models \cite{wolpert1998multiple,honda2018tandem}.\\
The aim of this work is to develop a controller based on a cerebellar-like model, developed on the cellular-level, to guide the motion of robots with real-time sensory information. The cerebellar model developed in this study is more detailed from the biological perspective to the previously developed model \cite{zahra2020vision,abadia2019robot} to demonstrate the effect of these additional features and its effect on the performance. The controller first builds a sensorimotor differential map through motor babbling and the cerebellum acts as a Smith predictor \cite{tolu2020cerebellum} to correct discrepancies in sensory readings to enhance accuracy and precision of robot movements.\\
While developing cellular-level model adds more challenges for tuning and constructing the network, it provides insight into the real working of the biological counterparts and allows to import additional features from the wide repository of studies exploiting neural mechanisms to achieve such features. Building a detailed cellular level model requires utilizing spiking neural network (SNN), the third generation of artificial neural networks (ANN), to give a more faithful representation of the neuronal dynamics which provide them with more complex and realistic firing patterns based on spikes \cite{maass1997networks}. The SNN adds a temporal dimension compared to the previous generations of ANN, which allows for developing more biologically realistic learning mechanisms and a more efficient representation of the spaces/dimensions encoded \cite{maass1997networks}.\\
In \cite{abadia2019robot}, a controller is developed based on an inverse cellular-level cerebellar model to enhance the robot movement. The controller relies on a trajectory planner and inverse kinematics model to generate reference signals for angular position and speed. This limits the ability to learn only the manipulation of the robot based on the given reference value and is not suitable to learn directly from sensory feedback. In \cite{Capoleietal2020}, a cerebellar model based on the adaptive filter theory is developed combining some computational neuroscience techniques along with  machine learning. Such combination aims to achieve the sensorimotor adaptation, but it does not model the spiking activity in the cerebellum. In \cite{zahra2020vision}, a cellular level cerebellar model is developed in which the teaching signal is provided based on the error in task-space.
In all the previous studies, no clear method was identified to tune the network parameters, and the developed networks were used to manipulate only the end-effector in some predefined trajectory or to reach a certain target.

This study contributes to building a detailed spiking cellular-level cerebellar model including more biological features compared to the previous studies. The parameters of the network are set using a Bayesian optimization method to preserve several biological features observed in the cerebellum. This is demonstrated by monitoring the activity and firing rates of the different groups of neurons in the cerebellum, and the output from each layer, which is then compared to those obtained from biological counterparts. The teaching signal is provided as the error in task-space and demonstrates the ability to adapt to executing different tasks and handling manipulation of deformable objects. \\
\\

\begin{figure}
\centering
\begin{subfigure}{0.49\columnwidth}
\centering
\includegraphics[width=\columnwidth]{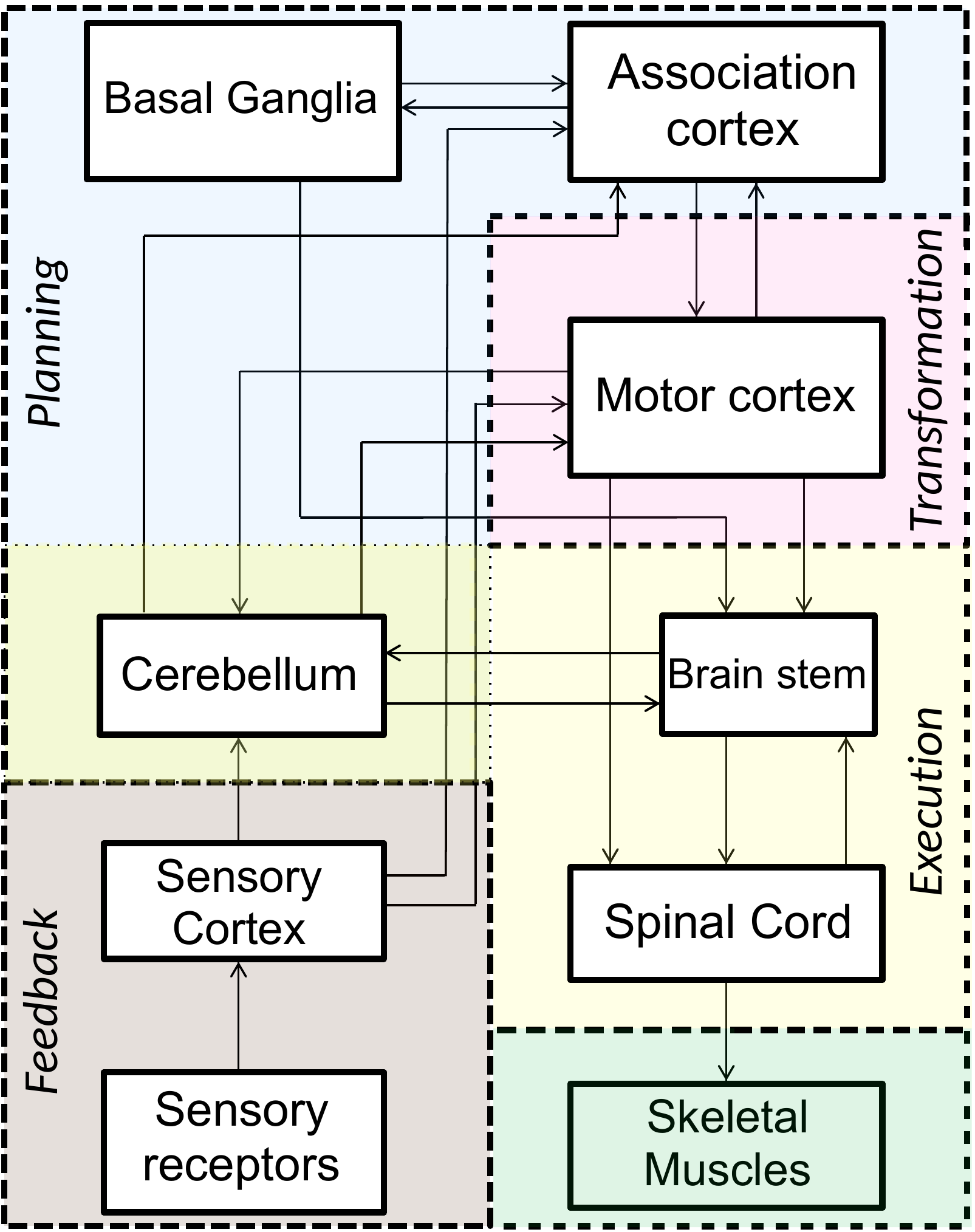}
\caption{}
\label{fig:motor_hierarachy}
\end{subfigure}%
\begin{subfigure}{0.49\columnwidth}
\centering
\includegraphics[width=\columnwidth]{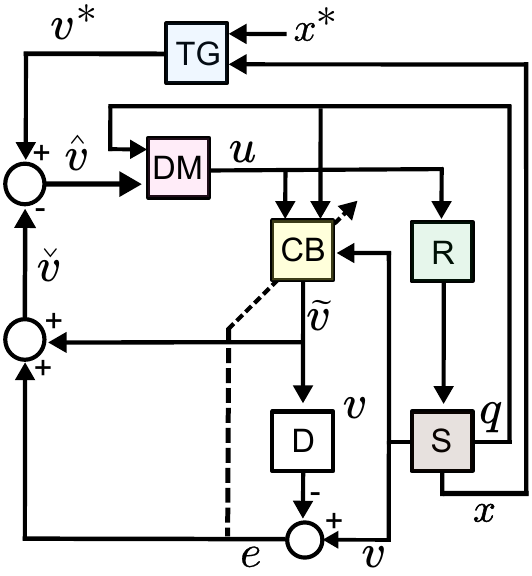}
\caption{}
\label{fig:smith_predictor}
\end{subfigure}
\caption{(a) A simplified schematic of the hierarchical motor system based on studies from the literature\cite{mogenson1980motivation,merel2019hierarchical}. The Thalamus is not included for clarity. (b) The block diagram for the proposed cerebellar-based control system for the robot $R$. Based on the motor command $u$ generated by the differential map $DM$, the forward cerebellar model $CB$ provides sensory predictions (i.e., the robot state) for the next cycle. Discrepancy between the actual state observed by the sensors $S$ and the predicted state is used to correct the desired state signal generated by the target generator $TG$ before introducing to $DM$.}
\label{fig:chart}
\end{figure}

In the next sections, the preliminaries are introduced (Sec. 2), then modeling of the cerebellar controller and the hyperparameter optimization method is discussed (Sec. 3). The experimental setup along with the results are explored (Sec. 4). Finally, an analysis of the obtained results is discussed and the main conclusions are derived (Sec. 5).

\section{PRELIMINARIES}\label{sec:pre}
The motor system is composed of several regions, each of which is responsible for a different function. These areas follow a hierarchical organization as shown in Fig.\ref{fig:motor_hierarachy}. In such hierarchy \cite{mogenson1980motivation,merel2019hierarchical}, both parallel and serial connections provide different behaviours in different situations (i.e., dependent upon sensory feedback).

The higher order areas are responsible for decision making and planning the sequence of motion while coordinating the activity of several limbs. Lower order areas, on the other hand, control muscles while regulating forces and velocities with changes in posture and various interactions with the environment. The higher level tasks start from the cortical association areas and prefrontal cortex (along with the Basal Ganglia), which receive sensory information (from the sensory cortex) to generate an abstract plan for motion and the sequence of execution. This plan is then transformed to motor commands in the motor cortex which send these commands to the brain stem and the cerebellum (as an efference copy). The cerebellum plays a role in the coordination of movements and adjustment of timing to attain fine movements. Signals from the motor cortex travel to trigger motor neurons which innervate the skeletal muscles. The motor neurons in the spinal cord control the limbs and the movement of the body, while those in the brain stem control facial and head movements.

In relation with the critical role of the cerebellum in both planning and execution of motion, this study focuses on the cerebellar corrections due to noisy sensory readings to obtain fine movements while executing the motor commands generated by the motor cortex to reach a target point in the space. Thus, a computational model of the cerebellum can help improve the performance of robotic controllers. In this work, a biologically inspired control system is built to guide a robot in a servoing task after performing motor babbling for several iterations, as shown in Fig.\ref{fig:chart}. Computational spiking models are developed to both form a coarse sensorimotor map through motor babbling, and to reproduce the cerebellum. The sensory input to the formed map is modulated through the cerebellum to enhance the movement and reduce the deviation from the desired path. The developed cerebellar controller is capable of guiding the robot based on real-time noisy data.

\section{METHODS}\label{sec:methods}

\subsection{The Cerebellar-Based Control Architecture}

\begin{figure}
	\centering
	\includegraphics[width=\columnwidth]{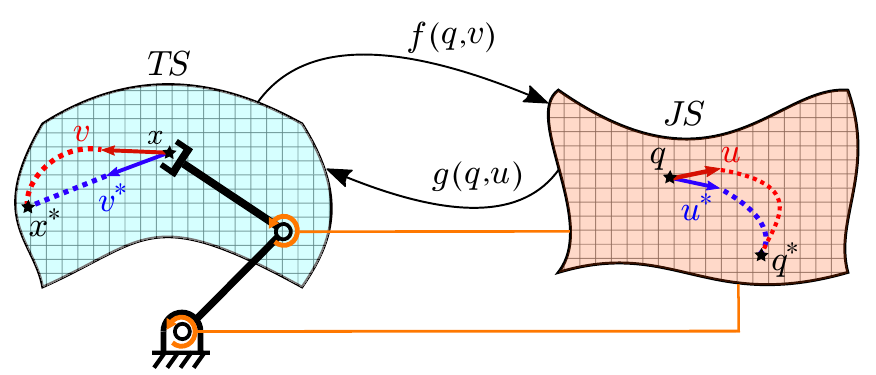}
	\caption{The forward and inverse models correlating task and joint spaces. In this study, $DM$ approximates the inverse model, while $CB$ approximates the forward model.}
\label{fig:space}
\end{figure}

As discussed earlier, various theories were developed about the formation of the cerebellar forward or inverse models. However, more biological evidences support the theory that the cerebellum acts as a forward model based on the behaviour observed in the case of cerebellar damage \cite{bastian2011moving}. In this study, the cerebellum acts as a computational forward model to predict the next sensory states based on the desired spatial velocity and the current robot states. This model fits in the designed controller to act as a Smith predictor which is known to be capable of handling control schemes with long dead time as shown in Fig.\ref{fig:chart}. Analogous to control systems, in which dead-time is introduced due to the time needed for sensing, processing of the inputs, computing the control output and actuation, in biological systems the dead-time is introduced due to the time needed for the sensory signals to travel through the nervous system, generate a motor command and travel back for execution to the muscles. In our system, the dead time (expressed as a delay ($D$)) is caused by slow sensory readings ($S$) and robot ($R$) action. In this context, the motor cortex acts as a differential map ($DM$) that can generate commands ($u$) in the joint space based on the desired spatial velocity $v^{*}$ and current joint angles $q$, and thus acts as an inverse model as shown in Fig.\ref{fig:space}.\\

 This can be expressed as:
\begin{equation}
    u(t) = f(q(t), v^{*}(t))
\end{equation}
where $f$ represents the inverse mapping formed in the motor cortex to correlate the joint space and task space. In case the robot moves from the current end-effector position $x$ to a target position $x^{*}$, $v^{*}$ can be formulated as:
\begin{equation}
    v^{*}(t) = \frac{x^{*}(t)-x(t)}{\|x^{*}(t)-x(t)\|}
\end{equation}
However, due to the delay $\tau$, discussed earlier, along with the imperfection of the map $f$, the motor command needs a correction before introducing it to the robot. Thus, the cerebellum acts as a forward model to predict spatial velocities upon applying $u$. While the robot motion can be expressed as:
\begin{equation}
    v(t)= g(q(t), u(t-\tau))
\end{equation}
where the map $g$ represents the actual differential kinematics of the robot, and $v(t)$ is the spatial velocity of the end-effector upon executing the command $u(t-\tau)$ while the robot joint configuration is $q(t)$. The spatial velocity predictions generated by the cerebellum $\tilde{v}(t)$ can then be expressed as:  
\begin{equation}
    \tilde{v}(t) = \tilde{g}(q(t), u(t-\tau), v(t),v^{*}(t))
\end{equation}
where $\tilde{g}$ represents the forward model built by the cerebellum to approximate the map $g$. The model approximations are improved during training relying on the feedback error $e$, where $e(t) = v(t) - \tilde{v}(t)$.
Hence, the error in sensory signals and discrepancy from the predicted value is used to modulate the desired spatial velocity before introducing to the $DM$, as detailed later in subsection \ref{sec:cereb}. So, $DM$ makes use of the predictions provided by the cerebellum to correct the anticipated error in sensory readings due to the dead-time effect:
\begin{equation}
    u(t) = f(q(t), v^{*}(t), \tilde{v}(t), \tilde{v}(t+\tau))
\end{equation}
Taking into consideration both the error from previous trials and the sensory prediction, the controller is enabled to correct the next motor command in an indirect way. This is carried out by adding the error from previous attempts to the predictions of the robot state to make up for the expected error:
\begin{equation}
   \Check{v}(t) =\tilde{v}(t+\tau) + e(t)  
\end{equation}
Finally, the corrected prediction $\Check{v}(t)$ is compared to the desired velocity $v^{*}$:
\begin{equation}
    \hat{v}(t) = 2v^{*}(t) - \Check{v}(t)
\end{equation}
To put it in other words, the $\hat{v}$ is the sensory signal that can be introduced to the $DM$ to give better estimations and make up for the delay in the feedback cycle.
\begin{equation}
    u(t) = f(q(t), \hat{v}(t))
\end{equation}

\subsection{The Differential Mapping SNN}

A two layer SNN, one input and one output layer, are connected via all-to-all plastic synapses (weights can change) to provide the transformation between two correlated spaces. The network encodes the current joint angles and the spatial velocity of the end-effector in the input layer and encodes the angular velocity of the joints in the output layer. As the network correlates mainly the velocities in two spaces, it acts as a differential map; it is named as Differential Mapping Spiking Neural Network (DMSNN) \cite{zahra2020differential}. For a robotic manipulator of $n$ degrees of freedom (DOF), the task space is represented by $n$ assemblies of neurons, where each dimension is encoded by a corresponding assembly.

\begin{figure}
\centering
\includegraphics[width=\columnwidth]{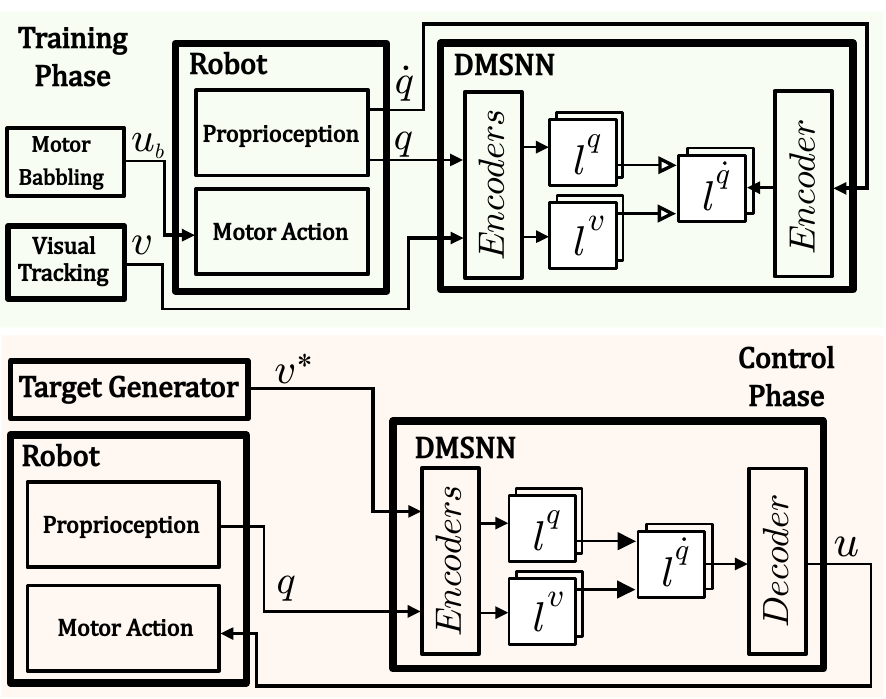}
\caption{Input (sensory) neurons are connected to output (motor) neurons through excitatory and inhibitory plastic synapses. The signals introduced to each neuron assembly are depicted. The motion is guided during the training of the network through motor babbling actions in joint space. After the training phase, the robot is controlled by decoding motor commands from the activity in the output layer. The hollow arrow heads refer to plastic synapses.}
\label{fig:dmsnn}
\end{figure}

Similarly, for $m$ DOF of joint space $m$ assemblies are needed. Hence, the input layer is made up of $n+m$ assemblies. The assemblies $l^{v_{1:n}}$ encode the $n$-dimensional spatial velocity, while the assemblies $l^{q_{1:m}}$ encode the $m$-dimensional joints' angular positions. At the output, the assemblies $l^{\dot{q}_{1:m}}$ encode the  $m$-dimesnional joints' angular velocities as depicted in Fig. \ref{fig:dmsnn}.
The two layers of the network are connected via all-to-all plastic synapses (both excitatory and inhibitory). The inhibitory synapses regulate the motor(output) neurons activity to maintain stable learning and hence avoid the unbounded increase in the weights of the excitatory synapses. At the output layer, local inter-inhibitory connections (i.e., within each assembly) are added with low inhibition to proximal neurons and higher inhibition to the distal neurons.

Modulation of the plastic synapses occurs during training the network as shown in Fig.\ref{fig:dmsnn}, where random target angles $q_d$ are generated for motor babbling. The robot joints move towards $q_d$ linearly in joint-space based on the error calculated from the difference between the desired joint angles and current one $q$. The internal (i.e.,\emph{proprioception}) and external sensors (i.e.,\emph{exterioception}) provide the necessary data to both the sensory and motor layers while training $DM$. These variables are encoded based on the preferred/central value $\psi_{c}$ defined for each neuron, and a distribution that allows all the neurons in the assembly to contribute to what is known as \emph{population coding} \cite{amari2003handbook}).
The neurons' firing rates are defined by the Gaussian tuning curve which can be expressed as:
\begin{equation}
\Theta_{i}(t) = \exp {\left(\dfrac{-\Vert{\psi -\psi_{c}  \Vert^{2}}}{2\sigma^{2}}\right)}
\end{equation}
where $\psi$ is the variable's value, and $\sigma$ is the radius calculated based on the number of neurons per assembly $N_{l}$, and the defined range of values of each variable.
The synaptic weights are modulated accordingly forming a proper $DM$. After training ends, the control phase starts where $DM$ is ready to to execute a coarse robotic servoing task. The corresponding values of the variable are encoded and introduced to assemblies $l^{q_{1:m}}$ and $l^{v_{1:n}}$, and the output is then decoded from the activity of $l^{\dot{q}_{1:m}}$. The decoding scheme in this case is the \emph{central neuron}  \cite{amari2003handbook}:
\begin{equation}
\psi_{est}= \dfrac{\Sigma \psi_{i}.\Theta_{i}}{\Sigma \Theta_{i}}
\end{equation}
where $\psi_{i}$ is the defined central value of neuron $i$ in $l^{\Psi}$ assembly $l^{\Psi}$, and $\psi_{est}$ is the decoded/estimated output value .
However, $DM$ is formed as a coarse map which lacks in both the precision and accuracy needed for fine control of the robot.

\subsection{The Proposed Forward Cerebellar-like Model}
\label{sec:cereb}

The cerebellum is composed of three layers. The \emph{Granule layer}, which is the innermost layer, is made up mainly of granule cells, which counts up to 80\% of the neurons in the brain \cite{herculano2009human}. It contains as well the \emph{Mossy fiber} axons and the Golgi cells. The \emph{Purkinje layer} is the middle layer and contains the Purkinje cells which are featured by the distinctive firing pattern and considered a key component for learning to occur in the cerebellum. The \emph{Molecular layer} is the outermost layer containing the axons extended from the granule cells to purkinje cells (known as parallel fibers) intersecting with axons extended from the \emph{inferior olive} (known as\\

\begin{figure}
\centering
\begin{subfigure}{\columnwidth}
\centering
\includegraphics[width=0.9\linewidth]{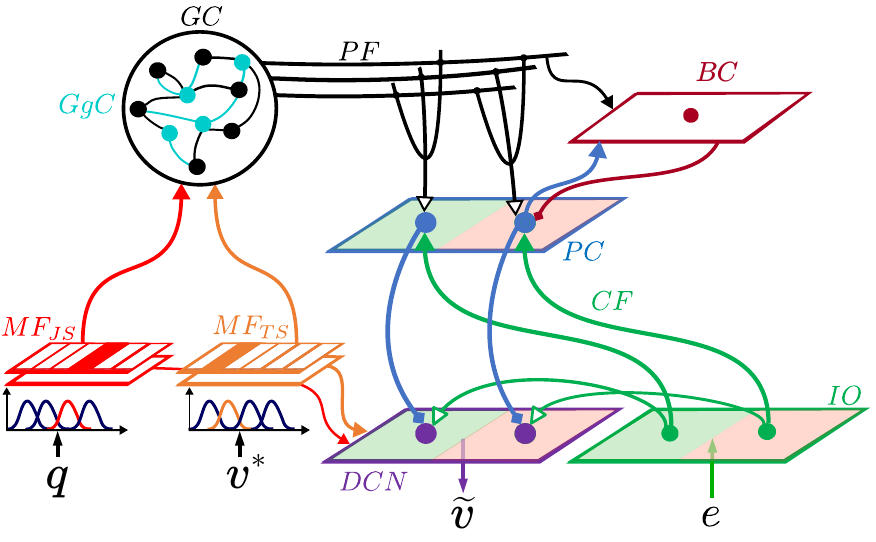} 
\caption{}
\label{fig:cereb_c}
\end{subfigure}
\begin{subfigure}{\columnwidth}
\centering
\includegraphics[width=\linewidth]{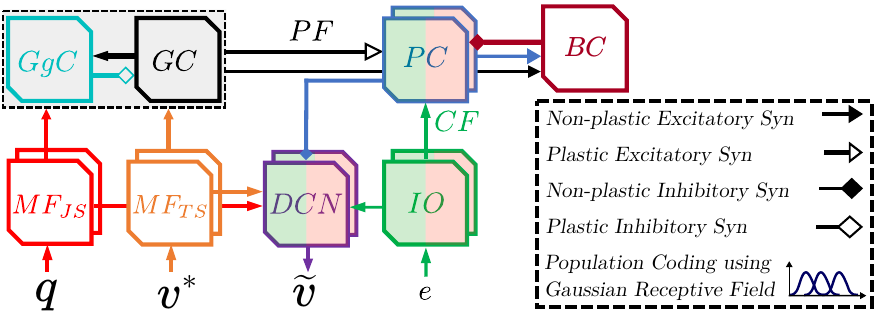} 
\caption{}
\label{fig:cereb_f}
\end{subfigure}
\caption{(a) A cellular and (b) functional schematic diagrams of the cerebellum.}
\label{fig:cereb}
\end{figure} 
\noindent climbing fibers). The molecular layer also contains the basket and stellate cells.

The cerebellar computational model developed, as shown in Fig. \ref{fig:cereb}, acts to rectify the sensory readings before introducing to the differential map ($DM$) which replicates the function of the motor cortex. This enhances the motor commands generated by $DM$ by accounting for sensory discrepancies in the previous cycles as explained in the two previous subsections. The cerebellar forward model is developed based on the cerebellar microcircuit. 

The \emph{Mossy Fibers} ($MF$) encode the current angular position of the joints and the desired Cartesian velocity. These two variables are chosen to provide the essential information to define the state of the robot in both joint space and task space. Thus, $MF$ consists of $n_{MF}$ assemblies of neurons, and each assembly encodes a dimension of one of the variables. The angular position is encoded by $n_{JS}$ assemblies forming the $MF_{JS}$ group, where $n_{JS}$ is the number of degrees of freedom (DOF) studied. The desired Cartesian velocity is encoded by $n_{TS}$ assemblies forming the $MF_{TS}$ group, where $n_{TS}$ is the number of Cartesian DOF studied. 

The synapses connecting between $MF$, Granular cells ($GC$) and Golgi cells ($GgC$) shall lead to a sparse coding of robot states. The Inferior olive ($IO$) provides the teaching signal, which in this study is the task space error, through the climbing fibers ($CF$) to Purkinje Cells ($PC$). The error ($e$) is defined as the discrepancy between the actual and the predicted spatial velocities. The plastic synapses connecting $GC$ to $PC$, known as parallel fibers $PF$, are modulated initially under the effect of the teaching signals from $CF$. These signals evoke activity in $PC$ to provide the desired correction, and thus allows to encode such corrections at the corresponding state of the robot. 

The $MF$ connects to Deep Cerebellar Nuclei ($DCN$) through excitatory synapses to maintain a basal spiking activity. The $PC$ connects to $DCN$ through inhibitory synapses to allow for the right value to be decoded from the activity of these neurons, while $IO$ connects to $DCN$ through excitatory plastic connections studied to provide fast convergence of learning \cite{luque2014fast}.  

 The $PC$,and similarly $IO$ and $DCN$, consists of $n_{TS}$ neurons' groups with each group consisting of two assemblies for positive and negative change for each DOF.

After the $DM$ training goes for several iteration ,i.e., till the coarse control map is formed, the training then starts for the cerebellum to build the corrective mapping. 

In \cite{nishiyori2016developmental}, a study was conducted on infants between 6 to 12 months to monitor the neural activity in the motor cortex (M1) while reaching targets. It was observed that the activity shifts from a diffused state across M1 to a focused one as the age of infants increases. Additionally, insufficient data is reported about the development of the cerebellum in infants at that age\cite{chugani2018imaging}. Thus, it is safe to assume that in infants the development of the cerebellum starts later than the motor cortex and its contribution increases with the increase in its size which is reflected by performing fine movements and exhibiting some motor skills \cite{knickmeyer2008structural}.

Similar to encoding in $DM$, the input values to $MF$ is first encoded employing population coding, with the current introduced to the $i^{th}$ neuronal unit in the $j^{th}$ $MF$ assembly ($\Theta_{MF_{i,j}}(t)$) can be calculated using the following equation:
\begin{equation}
\Theta^{MF}_{i,j}(t) = \exp {\left(\dfrac{-\Vert{\theta^{MF}_{j} -\theta^{MF}_{i,j}  \Vert^{2}}}{2\sigma_{MF_j}^{2}}\right)}
\end{equation}
where $\theta^{MF}_{j}$ is the input to the $j^{th}$ $MF$ assembly, and $\sigma_{MF_j}$ is the radius for the Gaussian distribution, and the variable ranges from $\theta_{MF_{j_{min}}}$ to $\theta_{MF_{j_{max}}}$. $\theta_{MF_{i,j}}$ is the preferred/central value of $i^{th}$ neuron in $j^{th}$ $MF$ assembly. Both $\theta_{MF_{i,j}}$ and $\sigma_{MF_j}$ are adjusted using the \emph{self-organization} algorithm (SOA), where the data previously collected for babbling $\Xi$ is used to adjust the values of $\theta_{MF_{i,j}}$ for all neurons.
\textbf{SOA:} A best matching unit $\beta$ is chosen for each central value from the linearly initialized set $\theta_{MF_{j}}$ (i.e., equally spaced from $\theta_{MF_{j_{min}}}$ to $\theta_{MF_{j_{max}}}$). $\beta$ is picked based on the Euclidean distance from a random data sample $\xi$ (from the set $\Xi$):
\begin{equation}
    \beta = \argmin{\kappa}{(\theta_{MF_{\kappa,j}} - \xi)^2}
\end{equation}
The value of $\theta_{MF_{\beta,j}}$ and that of the neighbouring units at an instant $k$ are updated such that:
\begin{align}
\begin{split}
    \theta_{MF_{i,j}}(k+1) = \theta_{MF_{i,j}}(k) +\\ \rho(k)\nu_{i\beta}(k)(\xi - \theta_{MF_{i,j}}(k))
\end{split}
\end{align}
where $\rho$ is the learning rate and $\nu$ is the neighbouring/proximity function given by:
\begin{equation}
    \nu_{i\beta}(k) = \exp\left(\frac{-\|i-\beta\|^2}{2\vartheta^2(k)}\right)    
\end{equation}
and $\vartheta$ is the radius gauging the proximity of the neighborhood.
Both $\rho$ and $\vartheta$ decay exponentially over the whole period $K$:
\begin{equation}
    \rho(k) = \rho_\circ \exp(\frac{-k}{K}),\vartheta(k)=\vartheta_\circ \exp(\frac{-k}{K})
\end{equation}
where $\rho_\circ$ and $\vartheta_\circ$ are the initial values for $\rho$ and $\vartheta$, respectively. After running the SOA for $K$ iterations, the radius $\sigma_{MF_{i,j}}$ is set for every neuron in $MF$ separately such that:
\begin{equation}
    \sigma_{MF_{i,j}} = \sigma_{MF_{i,j}} - \sigma_{MF_{i+1,j}}
\end{equation}
with the last radius (i.e., $\sigma_{MF_{N_l,j}}$) set to have the same radius as the previous one (i.e., $\sigma_{MF_{N_l-1,j}}$).

The application of the SOA to $MF$ allows for a more distinctive input to the $GC$ and better encoding of the robot states. Both $GC$ and $MF$ are connected through excitatory synapses to the Golgi Cells $GgC$, which is connected to the $GC$ through plastic inhibitory connections. The recurrence through the reciprocal connections between $GC$ and $GgC$ along with the connections to the $PC$ allows for spare encoding of the robot state space and can interpreted as a Liquid State Machine (LSM) as argued in \cite{yamazaki2007cerebellum}. It is also claimed that the inhibitory action of $GgC$ allows to have a minimum number of neurons in $GC$ active at the same time and thus higher sparsity and better encoding of the states \cite{kandel2000principles}.

Each neuron from $GC$ is connected to a randomly picked neuron from each assembly $MF$,and hence each neuron from $GC$ shall be connected to $n_{MF}$ neurons.
Furthermore, $MF$ connects via excitatory synapses (all-to-all connections) to $DCN$. $GC$ connects to $PC$ via plastic excitatory projections (which are known as Parallel Fibers $PF$). The parameters of the network shall be tuned such that $PF$ are modulated only when $IO$ neurons are active. The projections from $IO$ to $PC$, known as Climbing Fibers $CF$, are one-to-one synapses to ensure that neurons belonging to the same group (i.e., the same DOF) and direction (i.e., positive/negative changes) connect to each other, and hence ensuring that the right members of $CF$ are modulated. $IO$ connects through excitatory synaptic connections to $DCN$. The activity of neurons in $IO$ is given by:
\begin{align} 
        \Theta_{IO+_j} = 
        &\left\{
            \begin{array}{ll} 
          \Theta_{IO_{max}} & e_{pred} > \Upsilon \\
          0 & e_{pred}\leq \Upsilon
       \end{array}
        \right.  
        \\
        \Theta_{IO-_j} = 
        &\left\{
            \begin{array}{ll} 
          0 & e_{pred}\ge -\Upsilon \\
          \Theta_{IO_{max}} & e_{pred} < -\Upsilon
        \end{array}
        \right.
\end{align}

where $\Theta_{IO+_j}$ and $\Theta_{IO-_j}$ describe the mean firing rates of the two opposite directions of the $j^{th}$ DOF encoded by the $IO$ assemblies, while $\Theta_{IO_{max}}$ describes the maximum firing rate of neurons in $IO$. A threshold value $\Upsilon$ is defined to avoid an overlapping oscillatory activity around the reference value. 

Similarly, the connections between PC and DCN follow the same concept to allow for activation of the right group of neurons. The cerebellar output is decoded from the activity of $DCN$ neurons:

\begin{equation}
    \tilde{v}_j = \frac {\Sigma \Theta_{DCN+_{i,j}} - \Sigma \Theta_{DCN-_{i,j}}}{{\Theta_{DCN_{max}}} * n_{DCN}} *v_{max_j}
\end{equation}

where $\tilde{v}_j$ gives the anticipated/predicted velocity for the $j^{th}$ DOF, $\Theta_{DCN+_i,j}$ and $\Theta_{DCN-_i,j}$ are the firing rates of the $i_{th}$ neuron in the two opposing directions assemblies of the $j^{th}$ DOF, and $n_{DCN}$ is the number of neurons in each $DCN$ assembly. $\Theta_{DCN_{max}}$ is the maximal firing rate observed in $DCN$. $v_{max_j}$ defines the maximum speed for the $j^{th}$ DOF.

\subsection{Optimization of The Network Parameters}\label{sec:optimization}

To have the cerebellar microcircuits employ more features of the biological counterparts, an optimization process is applied to finely tune the parameters. The optimization allows to set the values of the  hyperparameters $\mathcal{H}$ of the network to meet specific goals for that mean. Such goals are set to tune the layer by layer of the cerebellar model through the defined objective function $f_i(\mathcal{H}_i)$ to be minimized, where $\mathcal{H}_i \subset \mathcal{H}$. Each objective function consists of a set of objectives $\mathcal{O}_i$ weighed by a vector $w_\mathcal{O}^i$, such that $f_i(\mathcal{H}_i)=\mathcal{O}_i w_\mathcal{O}^i$. Tuning the model layer by layer allows to properly define and monitor the expected output from each layer. Moreover, this allows to simplify the optimization and avoid the complexities accompanying choosing a large number of hyperparameters to be optimized simultaneously.

\textit{Objective 1: }\textbf{(Uniform firing in $MF$)}
The input signals coming from the $DM$ is first introduced to the $MF$. Thus, the first step is to make sure that the firing pattern in $MF$ is suitable to be introduced to the next layer. The parameters affecting the firing are the neuron parameters and the amplitude of the input current to the neurons. The criteria selected are the maximum firing rate of the neurons and the pattern of firing. For the Gaussian distribution of the input current, it is expected to have a corresponding Gaussian distribution, as well, for the firing rates in the $MF$ layer. Thus, a Gaussian distribution fitting, via maximum likelihood estimation \cite{cousineau2004fitting}, is applied to the number of spikes released from the neurons to compare the mean value of the curve $G_{mean}$ to the expected output (i.e., the neuron $MF_{nearest}$ whose central value is the closest to the input value). Hence, the two objectives are defined $\mathcal{O}_1 = [\mathcal{O}^1_1, \mathcal{O}^2_1]$ and can be formulated as:
\begin{align}
\begin{split}
 \mathcal{O}^1_1 = 
\frac{\sum_{n=1}^{N_{test}} |fr^{MF}_{max}(n)-fr^{MF}_{desired}|}{N_{test}}\\
\mathcal{O}^2_1 = \frac{\sum_{n=1}^{N_{test}}|G_{mean}(n)-MF_{nearest}(n)|}{N_{test}}   
\end{split}    
\end{align}

With $w_\mathcal{O}^1=[0.5,0.5]$, the first objective function is defined as $f_1(\mathcal{H}_1) = \mathcal{O}_1 w_\mathcal{O}^1$.

\textit{Objective 2: }\textbf{(Sparsity in $GC$)}
The output from $MF$ is then introduced to the $GC$. As mentioned in the previous subsection, $GC$ and $GgC$ act together to give a distinctive output for each input from the $MF$, and thus allowing for sparse coding of the input. Consequently, $f_2$ is designed to ensure that a minimum number of $GC$ neurons is active and checks as well for the uniqueness of  the activity pattern for every input. Additionally, the firing rate is defined to resemble the neuron activity in the biological counterpart. With these four objectives to be satisfied, $\mathcal{O}_2 = [\mathcal{O}^1_2, \mathcal{O}^2_2,\mathcal{O}^3_2,\mathcal{O}^4_2]$, the firing rates is defined in a way similar to that in $MF$, where:
\begin{align}
\begin{split}
 \mathcal{O}^1_2 = 
\frac{\sum_{n=1}^{N_{test}} |fr^{GC}_{max}(n)-fr^{GC}_{desired}|}{N_{test}}\\
\mathcal{O}^2_2 = 
\frac{\sum_{n=1}^{N_{test}} |fr^{GgC}_{max}(n)-fr^{GgC}_{desired}|}{N_{test}}
\end{split}    
\end{align}
The objective $\mathcal{O}^3_2$ targets minimizing the number of active neurons as much as possible, but also ensures that there is still active neurons in $GC$ (i.e., at least one neuron is active). During each iteration, the number of spikes triggered by each neuron is recorded in the set $S^{GC}_n$ after the $n^{th}$ test trial, to be used to define the firing rates and the active neurons as well. Neurons with the a firing rate greater than third of the desired firing rate (i.e., firing rate greater than $0.33fr^{GC}_{desired}$) are kept and the rest are discarded, then $S^{GC}_n$ is updated accordingly. This allows to consider only the neurons that would affect the learning in $PC$. The difference between the optimal number of firing neurons (chosen as 1 in this case) and the number of active neurons $\lambda^{GC}_n$ is recorded for each trial $n$ in a set $\Lambda^{GC}_n$. To penalize the state in which all neurons in $GC$ are inactive, a large number/score $\phi$ is returned, which is formulated as follows:
\begin{align}
\begin{split}
 \Lambda^{GC}_n = 
    \begin{cases} 
      \phi & \lambda^{GC}_n =0 \\
      |\lambda^{GC}_n-1| & \lambda^{GC}_n > 0 
   \end{cases}
\end{split}    
\end{align}
\begin{equation}
     \mathcal{O}^3_2 = \frac{\sum_{n=1}^{N_{test}} |\Lambda^{GC}_n-1|}{N_{test}}
\end{equation}

The objective $\mathcal{O}^4_2$ describes the uniqueness of the output obtained across the $N_{test}$ trials, where the repetition of spiking of a neuron across many trials is penalized. To simplify the computations, from each trial $n$ the neuron with maximum firing rate is recorded, and then, the number of repetitions of each of the neurons in the set is computed. The mean $\mu_{mean}$ and maximum $\mu_{max}$ number of repetitions is finally computed to formulate $\mathcal{O}^4_2$ as:
\begin{equation}
    \mathcal{O}^4_2 = 0.3\mu_{max} + 0.7\mu_{mean}
\end{equation}
With $w_\mathcal{O}^2=[0.1,0.1,0.2,0.6]$, the second objective function is defined as $f_2(\mathcal{H}_2) = \mathcal{O}_2 w_\mathcal{O}^2$.

\textit{Objective 3: }\textbf{(Proper firing rates in $PC$)}
The neurons of $PC$ are known to have distinguishable firing patterns
in reponse to different inputs. The input from $GC$ tends to trigger simple spikes ($SS$), while the input from $IO$/$CF$ triggers complex spikes ($CS$) in $PC$. These two spiking patterns differ from each other in many aspects, but in this study only their respective firing rates are considered. Moreover, the assemblies of neurons within each group tend to have an alternating activity for different directions of motion. Thus, three objectives are defined, $\mathcal{O}_3 = [\mathcal{O}^1_3,\mathcal{O}^2_3,\mathcal{O}^3_3]$. The first two objectives can be formulated as:
\begin{align}
\begin{split}
 \mathcal{O}^1_3 = 
\frac{\sum_{n=1}^{N_{test}} |fr^{PC}_{SS}(n)-fr^{SS}|}{N_{test}}\\
\mathcal{O}^2_3 = \frac{\sum_{n=1}^{N_{test}}|fr^{PC}_{CS}(n)-fr^{CS}|}{N_{test}}   
\end{split}    
\end{align} 
To construct the formula to describe the third objective $\mathcal{O}^3_3$, firstly the activity of the $PC$ is compared to a threshold value (chosen as the maximum firing frequency of the simple spikes) with those above and below the threshold assigned as active/true and inactive/false, respectively. In this case, the most desirable state is to only have one group of neurons active per DOF, which is represented by an XNOR logic gate:

\begin{align}
 A(PC_{\pm j}) =
 \begin{cases}
   1 & fr^{PC_{\pm j}} > fr^{PC}_{SS}\\
   0 & fr^{PC_{\pm j}} \leq fr^{PC}_{SS}
 \end{cases}
\end{align} 
where $fr^{PC_{\pm j}}$ is the mean firing rate of neuron in either of the two assemblies of neurons for each DOF $j$. $\mathcal{O}^3_3$ can then be expressed as:
\begin{equation}
 \mathcal{O}^3_3 = \frac{1}{n_{TS}}\sum_{j=1}^{n_{TS}}A(PC_{+j})\odot A(PC_{-j})  
\end{equation}
 
With $w_\mathcal{O}^3=[0.2,0.2,0.6]$, the third objective function is defined as $f_3(\mathcal{H}_3) = \mathcal{O}_3 w_\mathcal{O}^3$.

\textit{Objective 4: }\textbf{(Proper output from $DC$)}
After optimization is carried for the previous hyperparameters, this last objective function is directed to test the operation of the network while optimizing the parameters affecting the activity in the $DCN$. The robot, in a simulation environment, repeats a chosen motion towards a target four times, and both the error while moving $e_{pred}$ and the execution time $\Delta$ are recorded. The error in this case is formulated as:
\begin{equation}
        e_{pred} = \left|\arccos{\left(\frac{\Vec{\tilde{v}}\boldsymbol{\cdot} \Vec{v}}{\| \Vec{\tilde{v}}\| \|\Vec{v}\|}\right)}\right|
\label{eq:e_pred}
\end{equation}
Consequently, $\mathcal{O}_4$ consists of six weighted objectives, where $\mathcal{O}_4 = [\mathcal{O}^1_4,\mathcal{O}^2_4,\mathcal{O}^3_4,\mathcal{O}^4_4,\mathcal{O}^5_4,\mathcal{O}^6_4]$. The objective $\mathcal{O}^1_4$ acts to keep the mean value of the error $e_{pred}$ across the four trials as minimum as possible, and an inverted firing pattern in the opposing groups of neurons within the $DCN$ and when compared to $PC$ as well. The objectives $\mathcal{O}^2_4$ and $\mathcal{O}^3_4$ are set to promote the decrease in the the error $e_{pred}$ and the execution time $\Delta$, respectively, as the training proceeds. To achieve this, each of these objectives is assigned a value of 1 at the beginning of the training, and the variables (mean value of $e_{pred}$ and $\Delta$) are compared to those from the previous trial, to deduct $0.33$ in case of a decrease in the value of the variable. Hence, in case of a consistent decrease in the error and execution time from one trial to another, reflecting a successful learning process, the values of these objectives would return a value zero, which is the absolute minimum in this case. The objective $\mathcal{O}^4_4$ adjusts the firing rate of $DCN$ neurons in the desired range such that
\begin{align}
\begin{split}
 \mathcal{O}^4_4 = 
\frac{\sum_{n=1}^{N_{test}} |fr^{DCN}_{max}(n)-fr^{DCN}_{desired}|}{N_{test}}
\end{split}    
\end{align}
Similar to $PC$, in $DCN$ the opposing groups of neurons shall be set to fire in an alternating manner, which is formulated as:
\begin{align}
 A(DCN_{\pm j}) =
 \begin{cases}
   1 & fr^{DCN_{\pm j}} > fr^{DCN}\\
   0 & fr^{DCN_{\pm j}} \leq fr^{DCN}
 \end{cases}
\end{align} 
where $fr^{DCN}$ is the mean firing rate of $DCN$, and $fr^{DCN_{\pm j}}$ is the mean firing rate of neuron in either of the two assemblies of neurons for each DOF $j$. $\mathcal{O}^5_4$ can then be expressed as:
\begin{equation}
 \mathcal{O}^5_4 = \frac{1}{n_{TS}}\sum_{j=1}^{n_{TS}}A(DCN_{+j})\odot A(DCN_{-j})  
\end{equation}
Additionally, the objective $\mathcal{O}^6_4$ ensures that the activity in the assemblies of $DCN$ opposes that of the corresponding ones for same DOF in $PC$ (i.e., fire in an inverted manner):
\begin{equation}
 \mathcal{O}^6_4 = \frac{1}{n_{TS}}\sum_{j=1}^{n_{TS}}A(DCN_{\pm j})\odot A(PC_{\pm j})  
\end{equation}
With $w_\mathcal{O}^4=[0.3,0.1,0.1,0.1,0.2,0.2]$, the fourth objective function is defined as $f_4(\mathcal{H}_4) = \mathcal{O}_4 w_\mathcal{O}^4$.

\subsection{Bayesian Optimization for The Objectives}\label{sec:optimization_bayes}

To meet these objectives, Bayesian Optimization (BO) is utilized to optimize the defined objective functions \cite{bergstra2011algorithms}. These functions would be very costly and time consuming to optimize through manual or random searching methods due to the high dimensionality and stochasticity of the search space. BO develops a probabilistic model for the objective function to facilitate the evaluation of the objective function while making use of the history of previous trials to guide the optimization process, as explained later in this subsection. Thus, the optimal solution is sought to minimize each objective function to obtain the optimal hyperparameters $h^*_i$, such that:
\begin{equation}
    h_i^* = \argmin_{h_i\in \mathcal{H}_i}f_i(h_i)
\end{equation}
The main constituents through which a BO method is identified are the regression model and the acquisition function. The probabilistic regression model \emph{surrogates} the objective function (and referred to usually as the surrogate model). This probabilistic model is initiated with some random evaluations to guide the algorithm, starting from complete uncertainty (prior), and develops as more evaluations are stored in the history to give better future evaluations and decrease the uncertainty (posterior). The surrogate model is expressed as $\mathcal{S}_i = P(\mathcal{L}_i|\mathcal{H}_i)$ representing the mapping of the $\mathcal{H}_i$ hyperparameters to a probability of a loss/score $\mathcal{L}_i$ for an objective function $f_i$. The acquisition function (also known as selection function) allows for selecting appropriate candidates to improve the surrogate model while exploring for the optimum values for the hyperparameters. Hence, a proper choice of the acquisition functions guarantees a balance between exploration and exploitation without getting trapped in a local minima.
In this study, an \emph{Adaptive Tree Parzen Estimator} (ATPE) is chosen as a regression model \cite{arsenault_2018}, and the \emph{Expected Improvement} (EI) as the acquisition function. 

The standard TPE is known to fit for optimization problems where either a mixture of discrete and continuous hyperparameter spaces are to be studied or when the hyperparameters are contingent upon each other, and it is chosen for the latter reason. In TPE, rather than describing the posterior $P(\mathcal{L}_i|\mathcal{H}_i)$, it describes instead $P(\mathcal{H}_i|\mathcal{L}_i)$, relying on Bayes rule such that:
\begin{equation}
   P(\mathcal{L}_i|\mathcal{H}_i) = \frac{P(\mathcal{H}_i|\mathcal{L}_i) P(\mathcal{L}_i)}{P(\mathcal{H}_i)}
\label{eq:bayes}
\end{equation}
The TPE targets building two separate hierarchical processes, $P(\mathcal{H}_i|\mathcal{L}_i \in \mathcal{U})$ and $P(\mathcal{H}_i|\mathcal{L}_i \in \mathcal{D})$, where the sets $\mathcal{U}$ and $\mathcal{D}$ contain the highest and lowest values of $\mathcal{L}_i$, respectively, observed so far reference to a defined threshold value $\mathcal{L}^*_i$. This threshold value is decided based on a predefined percentage $\gamma$ such that $P(\mathcal{L}_i<\mathcal{L}^*_i)=\gamma$. The likelihoods $\mathcal{U}$ and $\mathcal{D}$ are modelled via kernel density estimators (which in this case is the \emph{Parzen estimator}). The Parzen estimator PE allows to represent a function through a mixture of kernels $K$, which are continuous distributions, to be expressed as:
\begin{equation}
    P(\mathcal{H}) = \frac{1}{N_{p}\eta} \sum_{j=1}^{N_{p}} K\frac{\mathcal{H} - \mathcal{H}_j}{\eta}
\end{equation}
where $N_{p}$ is the number of kernels used for the approximation, $\eta$ is the bandwidth of each kernel, and $K$ is chosen to be a normal distribution. Modelling $\mathcal{U}$ and $\mathcal{D}$ gives a way to choose hyperparameters for the next observations that are more likely to return lower values for the objective functions (in the case of minimization of objective functions).

Although TPE has less time complexity compared to other BO methods (as Gaussian Process BO), TPE does not model interaction/correlations among the hyperparameters. The ATPE addresses this drawback by juding from Spearman correlation \cite{zar2005spearman} between the hyperparameters which parameters to vary and which parameters to lock to achieve a more efficient exploration.
Also, among the drawbacks of TPE is that it has a fixed value for $\gamma$ and a fixed number of candidates introduced to the acquisition function to predict the next candidate optimal solution, which were introduced initially while solving some specific problems \cite{bergstra2011algorithms}. ATPE introduced empirically concluded formulas based on the cardinality of the search spaces for the hyperparameters to give better values for these two variables.

 In this study, the \emph{Expected Improvement} EI is chosen as the acquisition function, to maximize the ratio $P(\mathcal{H}_i|\mathcal{L}_i \in \mathcal{D})/P(\mathcal{H}_i|\mathcal{L}_i \in \mathcal{U})$. The EI generates a probability of obtaining a better solution than the current optimum solution and the amount of expected improvement as well, and consequently, favors bigger improvements. The basic formula for the EI is \cite{bergstra2011algorithms}:
\begin{equation}
    EI_{\mathcal{L}^*_i}(\mathcal{H}_i) = \int_{-\infty}^{\mathcal{L}^*_i} (\mathcal{L}^*_i - \mathcal{L}_i)P(\mathcal{L}_i|\mathcal{H}_i) \,d\mathcal{L}_i 
\label{eq:ei}
\end{equation}
By applying Bayes rule (equation \ref{eq:bayes}) and substituting in equation \ref{eq:ei}, the EI can be written as \cite{bergstra2011algorithms}:
\begin{gather}
    EI_{\mathcal{L}^*_i}(\mathcal{H}_i) = \frac{\gamma \mathcal{L}^*_i \mathcal{D}(\mathcal{H}_i)-\mathcal{D}(\mathcal{H}_i)\int_{-\infty}^{\mathcal{L}^*_i}P(\mathcal{L}_i) \,d\mathcal{L}_i}{\gamma \mathcal{D}(\mathcal{H}_i)+(1-\gamma)\mathcal{U}(\mathcal{H}_i)}\\
    EI_{\mathcal{L}^*_i}(\mathcal{H}_i) \propto (\gamma + \frac{\mathcal{U}(\mathcal{H}_i)}{\mathcal{D}(\mathcal{H}_i)}(1-\gamma))^{-1}
\label{eq:ei_bayes}
\end{gather}
From the formula \ref{eq:ei_bayes}, it can be concluded that EI acts to maximize the ratio $\mathcal{D}(\mathcal{H}_i)/\mathcal{U}(\mathcal{H}_i)$ and thus leading to introducing better candidates for the next search while still maintaining a trade-off between exploration and exploitation.

\subsection{Neuron Model}
To model the spiking activity of the neurons in the cerebellum in real-time, the simple neuron model developed by Izhikevich is chosen for its ability to reproduce different firing patterns \cite{izhikevich2003simple}. The model provides a decent biological plausibility at a relatively low computational cost.
The following differential equations describe the model:
\begin{align}
\dot{\mathcal{V}} & = f(\mathcal{V},\mathcal{U}) = 0.04\mathcal{V}^2+5\mathcal{V}+140-\mathcal{U}+I    \label{eq:update_v} \\
\dot{\mathcal{U}} & = g(\mathcal{V},\mathcal{U}) = a(b\mathcal{V}-\mathcal{U})               \label{eq:update_u}
\end{align}
Additionally, the membrane potential is reset after triggering a spike such that:
\begin{equation}
\label{eq:reset}
\text{if }v\ge30 \text{ mV}, \quad
\text{then } v\leftarrow c,~ u\leftarrow (u+d)
\end{equation}
where $\mathcal{V}$ (in $mV$) is the membrane potential and $\mathcal{U}$ is the variable that acts to lower the neuron's membrane potential (which is also known as the recovery variable).
The parameter $a$ (in $ms^{-1}$) decides the time scale of $\mathcal{U}$ (i.e., the decay rate), and $b$ (dimensionless) describes the sensitivity of the neuron before trigerring a spike (i.e., sensitivity of $\mathcal{U}$ to the sub-threshold $\mathcal{V}$). The parameter $c$ (in $mV$) describes the reset value of $\mathcal{V}$ after a spike is triggered, and $d$ (in $mV$) gives the reset value of $u$ after triggering spike. The external currents introduced to the neurons are described by $I$.

\subsection{Synaptic Connections}
To model the plastic connections in both $DM$ and $CB$, the Spike-timing-dependent plasticity (STDP) learning rule is chosen. In STDP, the change in the synaptic weight depends on the spikes relative timing in both the pre and post synaptic neurons \cite{stdp}.
For the $CB$, the antisymmetric STDP \cite{stdp_asym_pic} modulates the weight of the plastic synapses, and is formulated as:
\begin{equation}
\Delta {\varepsilon}_{ij} = \left\{
        \begin{array}{ll}
            -S_{a} \exp \left( {-\Delta t}/{\tau_{a}} \right) & \quad \Delta t \leq 0 \\
            \\
             S_{b} \exp \left( {-\Delta t}/{\tau_{b}} \right) & \quad \Delta t > 0
        \end{array}
    \right.  
\label{eq:asym_STDP}
\end{equation}
where the two coefficients $S_{a}$ and $S_{b}$ decide the amount of the decrease and increase of the synaptic weight, respectively. $\tau_{a}$ and $\tau_{b}$ determine the time windows through which synaptic weights decrease and increase, respectively.

The symmetric STDP is applied for the plastic synapses in $DM$ \cite{woodin2003coincident}, and can be formulated as:
\begin{equation}
\Delta {\varepsilon}_{ij} = S \left( 1-\left( {\Delta t}/{\tau_{1}} \right)^2\right) \exp\left({|\Delta t|}/{\tau_{2} } \right)
\label{eq:sym_STDP}
\end{equation}
where $S$ determines the amount of change in synaptic weights, while the ratio between the $\tau_{1}$ and $\tau_{2}$ adjusts the time window for increasing and decreasing the synaptic weights. $\Delta t$ is the difference between the timing of spikes at post-synaptic and pre-synaptic neurons, respectively, such that $\Delta t = t_{post} - t_{pre}$ .

\section{RESULTS}\label{sec:method_ver}
\subsection{Setup}
\begin{figure}
\centering
\begin{subfigure}{0.465\columnwidth}
\centering
\includegraphics[width=\columnwidth]{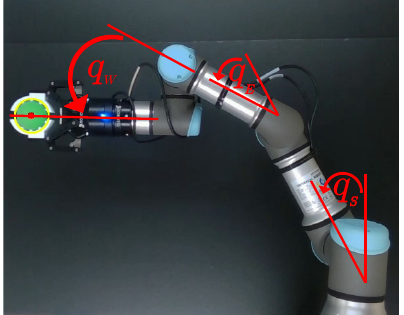}
\caption{}
\label{fig:robot_setup}
\end{subfigure}%
\begin{subfigure}{0.535\columnwidth}
\centering
\includegraphics[width=\columnwidth]{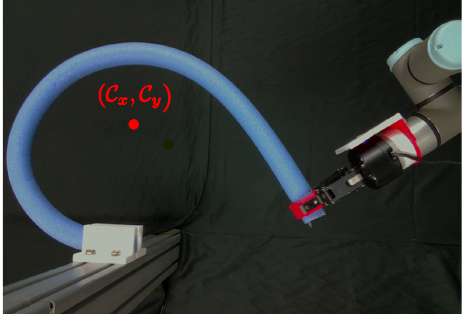}
\caption{}
\label{fig:robot_de_setup}
\end{subfigure}
\caption{The robot setups to manipulate (a) the end-effector, and (b) the deformable object.}
\label{fig:robot}
\end{figure}

A UR3 univeral robot is used to test the developed cerebellar controller in two experiments. The shoulder and elbow joints are controlled in an experiment to test the manipulation of the end-effector to a desired position as shown in Fig. \ref{fig:robot_setup}, while the elbow and wrist joints are controlled to test the manipulation of a deformable object as shown in Fig \ref{fig:robot_de_setup}. The shoulder and elbow joints for the first experiment are defined to have ranges of $q_S=[-170\degree,-135\degree]$ and $q_E=[-60\degree,0\degree]$, respectively. While the elbow and wrist joints for the first experiment are defined to have ranges of $q_E=[-45\degree,-20\degree]$ and $q_W=[-210\degree,-180\degree]$, respectively. A camera is fixed on top of the robot to track the end-effector position and the deformable object through color filtration. The proprioceptive readings (i.e, joints' positions and velocities) of the robot joints are obtained from motor encoders. The information required for motor babbling is recorded by giving commands to move to random joint angles linearly in joint-space within he defined ranges, through a script-based programming language developed for universal robots. \\
For the end-effector manipulation task, the robot is instructed to move to only 100 points in the joint space and the collected data during motion is then used to train $DM$, while for the deformable object case the robot is instructed to move to 300 points as small increments in joint angles may lead to a big change in the centroid of the deformable object and hence more rich data is needed. The spiking neural networks are developed using NeMo package \cite{nemo} which allows simulating the network using GPU. In this study, a computer with i7-6700K CPU and a GeForce GTX 1080Ti is used to build the networks and control the robot.

\section{Optimization Results}\label{sec:method_ver}
\begin{figure}[!b]
\centering
\begin{subfigure}{1.0\columnwidth}
\centering
\includegraphics[width=1.0\columnwidth]{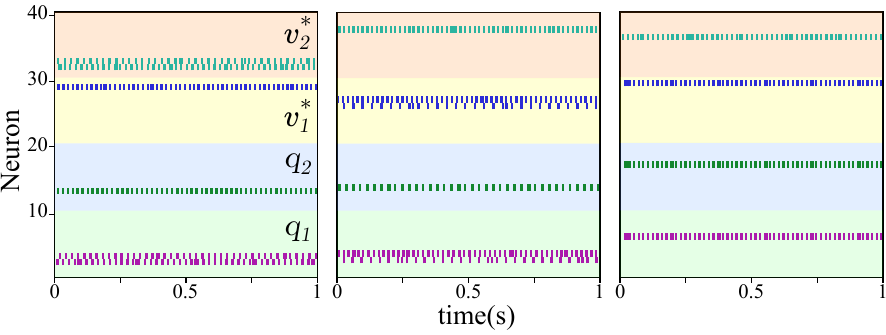} 
\caption{}
\label{fig:MF}
\end{subfigure}
\begin{subfigure}{1.0\columnwidth}
\centering
\includegraphics[width=1.0\columnwidth]{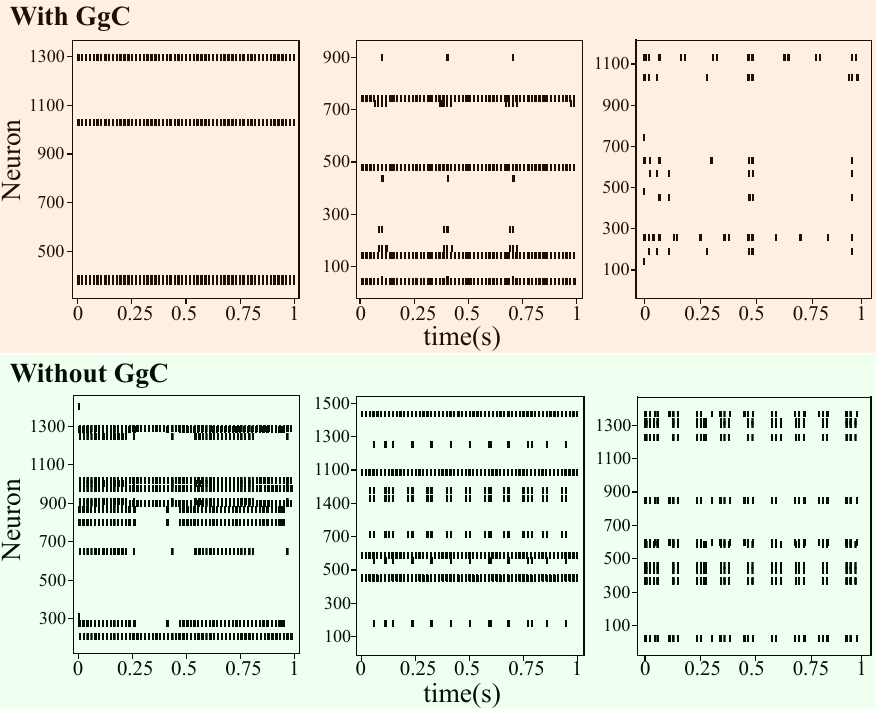}
\caption{}
\label{fig:GC}
\end{subfigure}
\caption{Firing in (a) $MF$ and (b) $GC$ with and without inhibition from $GgC$.}
\label{fig:MF_GC}
\end{figure}

After running the optimization described in the two previous subsections, the chosen parameters are obtained as shown in table \ref{table:cb_params}. The optimization is run on a robot in the simulation environment to avoid wear of the mechanical parts and for safety, but the final experiments are conducted on a real robot. Neurons in $MF$ achieve a maximum firing rate of 62 Hz which is comparable to 60$\pm$35 Hz spontaneous firing rate observed at excitatory synapses connecting $MF$ to $GC$, and an average firing rate (for active neurons only) of 40 Hz compared to 20 $\pm$ 21  \cite{powell2015synaptic}. The activity in each assembly is shown in Fig.\ref{fig:MF}. 

For the neurons in $GC$ the sparsity is achieved by having 6$\pm$3 neurons only active at the same time for a certain input, while maintaining a maximum firing rate of 86 Hz compared to 106 $\pm$ 65 Hz reported in \cite{powell2015synaptic} for maximal firing rate in $GC$ while at locomotion state. The sparse activity is achieved by the aid of firing in $GgC$ and the plasticity in synapses between $GC$ and $GgC$, to suppress the undesirable activity and thus limiting number of  active neurons. 
\begin{figure}
\centering
\begin{subfigure}{0.9\columnwidth}
\centering
\includegraphics[width=1.0\linewidth]{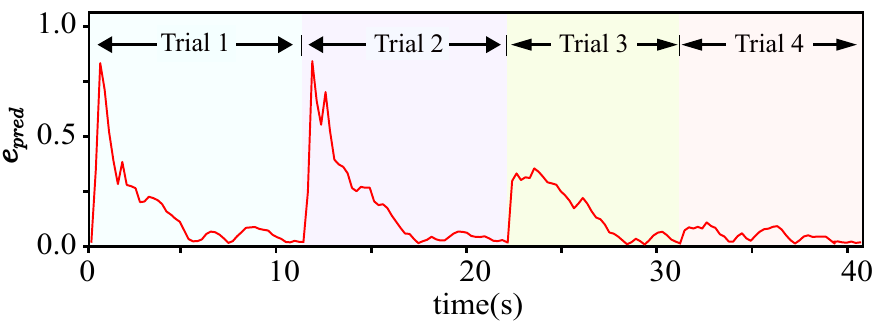} 
\caption{}
\label{fig:cb_cells}
\end{subfigure}
\begin{subfigure}{0.9\columnwidth}
\centering
\includegraphics[width=1.0\linewidth]{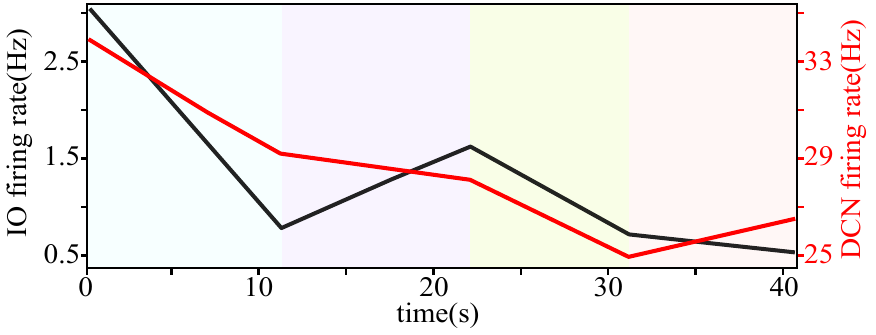}
\caption{}
\label{fig:DCN}
\end{subfigure}
\begin{subfigure}{0.5\columnwidth}
\centering
\includegraphics[width=1.0\linewidth]{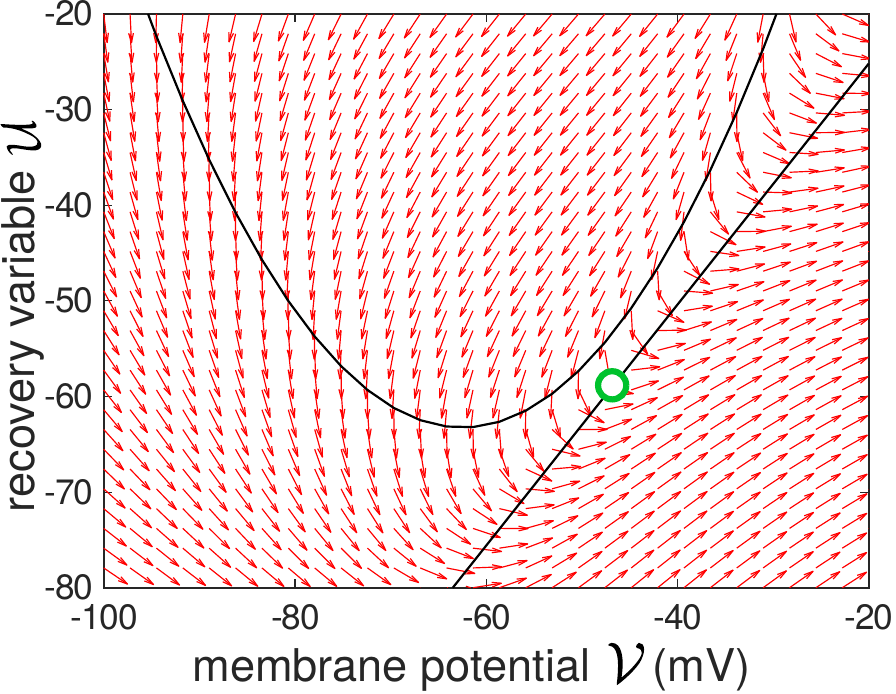} 
\caption{}
\label{fig:phase_portriat}
\end{subfigure}%
\begin{subfigure}{0.5\columnwidth}
\centering
\includegraphics[width=1.0\linewidth]{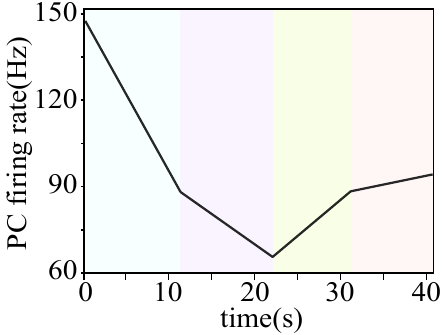}
\caption{}
\label{fig:PC}
\end{subfigure}
\caption{ (a) The error in cerebellar predictions $e_{pred}$, as defined in equation \ref{eq:e_pred}, as the learning proceeds for 4 trials, and (b) the average firing rates of $IO$ and $DCN$. (c)The phase portrait and (d) firing rates of $PC$.}
\label{fig:cereb}
\end{figure}

The parameters of $PC$ neurons are close to that of bistable neurons demonstrated by Izhikevich \cite{izhikevich2004model}, where analysis of the $PC$ properties in \cite{engbers2013bistability} provide indications of bistable behaviour. Hence, the neurons in $PC$ fire simple spikes with a firing rate of 70Hz (i.e., when no input is introduced from $IO$ through the $CF$) which is comparable to an average of firing rate of 50 Hz \cite{squire2012fundamental}. In case of complex spikes a firing rate of 160Hz is achieved , which is mentioned to achieve in biological counterpart to firing rates up to 400 Hz  \cite{squire2012fundamental}. While the modeled $PC$ fire slower than the real cells, however the big difference in the firing rates of simple and complex spikes facilitates the choice of adequate learning parameters for proper modulation of the $PF$ synapses. It shall be noted from Fig. \ref{fig:PC} that higher firing rates are achieved in the beginning, corresponding to bigger errors (and thus higher activity in $IO$), then the rate of activity decreases. After several trials, the strength of $PF$ synapses increases, and hence the firing rate increases. The spikes generated by $IO$ is characterized by a low rate ($<$3Hz) as seen in Fig. \ref{fig:DCN}, which is comparable to a firing rate 1 Hz reported \cite{mathews2012effects}, however a large number of $CF$ synapses connects between each neuron of $PC$ to a corresponding neuron of $IO$. This allows this low frequency stimulus to trigger high frequency spikes in $PF$. The maximum firing rate in $DCN$ is 40 Hz, which is comparable to the instantaneous firing rate 30 $\sim$ 50 Hz reported in \cite{lang2011control}.

\subsection{Radial Reaching}\label{sec:Sequence_servo}
In \cite{Fortier2002CEREBELLARAD}, patients with cerebellar damage attempt to reach radially towards targets at the same distance\\

\begin{table}
\caption{$CB$ Network Parameters\label{tb:cereb}}

\begin{tabular}{ |p{2.1cm}||p{0.62cm}|p{0.6cm}|p{0.6cm}|p{0.6cm}|p{0.6cm}|  }
 \hline
 \multicolumn{6}{|c|}{Values of neuronal parameters} \\
 \hline
 Area & $a$ & $b$ & $c$ & $d$ & $\mathcal{N}$ \\
 \hline
 $MF$ & 0.2 & 0.17 & -59. &14. & 40\\
 $GC$ & 0.22 & 0.25 & -55. &7. & 1500\\
 $GgC$ & 0.16 & 1.15 & -66. &16. & 7\\
 $PC$ & 1.74 & 1.24 & -59. &6. & 12\\
 $BC$ & 0.95 & 0.4 & -68. &16. & 70\\
 $IO$ & 0.02 & 0.25 & -65. &6. & 12\\
 $DCN$ & 0.45 & 0.08 & -56. &17. & 12\\
 \hline
\end{tabular}

\begin{tabular}{ |p{2.1cm}||p{0.9cm}|p{0.8cm}|p{0.9cm}|p{0.9cm}|  }
 \hline
 \multicolumn{5}{|c|}{Values of synaptic parameters} \\
 \hline
 Projection & T & V & $\omega_{init}$ & $\omega_{max}$\\
 \hline
 $MF\rightarrow GC$ & Rnd & 4 & 3.6 &-\\
 $MF\rightarrow GgC$ & Rnd & 1 & 0.6 &-\\
 $MF\rightarrow DCN$ & A2A & 1.0 & 9.0 &-\\
 $GC\rightarrow GgC$ & Prob & 0.01 & 0.3 &-\\
 $GgC\rightarrow GC$ & Prb & 0.5 & -9.8 &-15.0\\
 $GC\rightarrow PC$ & Prb & 0.8 & 0.01 &24.0\\
 $IO\rightarrow PC$ & O2O & - & 43.0 &-\\
 $IO\rightarrow DCN$ & A2A & 1.0 & 0.47 &-\\
 $PC\rightarrow DCN$ & O2O & - & -13.0 &-\\
 $PC\rightarrow BC$ & Prb & 0.4 & 2.4 &-\\
 $GC\rightarrow BC$ & Prb & 0.3 & 3.2 &-\\
 $BC\rightarrow PC$ & Prb & 0.5 & -45.4 &-\\
 \hline

\end{tabular}
\begin{tablenotes}
   \item[*] The 'T' column gives the connections' type from A to B in 'Projection' with a value 'V' for the parameter. 'Rnd' implies that 'V' neurons randomly picked from A connect to one neuron in B. 'Prb' implies that for each neuron in A there is a probability V to connect to each neuron in B. 'A2A' denotes All-to-All connections, where all neurons from A connect to all neurons in B. 'O2O' denotes One-to-One connections, where a neuron in A connects to a corresponding neuron in B. $\mathcal{N}$ is the total number of neurons of a specific type. $\omega_{init}$ and $\omega_{max}$ are the initial and maximum value of synaptic weights, respectively.
\end{tablenotes}
\label{table:cb_params}
\end{table}

\begin{figure}
\centering
\begin{subfigure}{\columnwidth}
\centering
\includegraphics[width=\columnwidth]{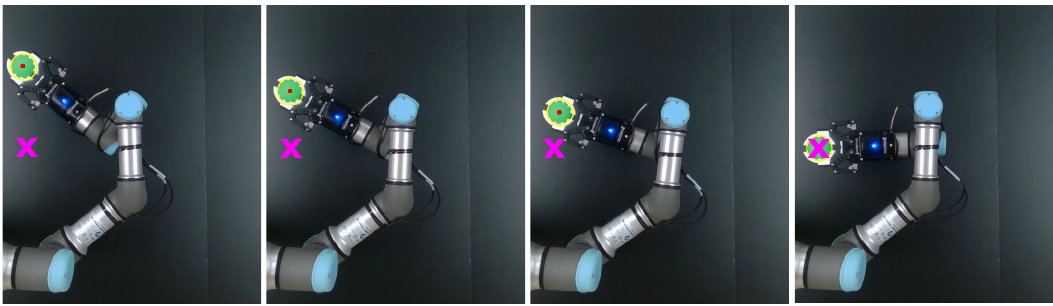}
\caption{}
\label{fig:rep_motion}
\end{subfigure}
\begin{subfigure}{\columnwidth}
\centering
\includegraphics[width=\columnwidth]{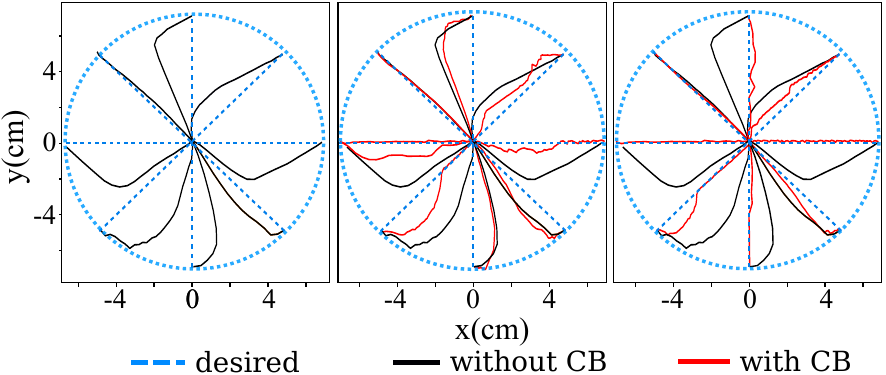}
\caption{}
\label{fig:star_servo}
\end{subfigure}

\caption{(a)A representative motion of target reaching. (b)The robot reaches the eight radial targets starting from the center of the drawn circle with targets apart 45$\degree$ from each other. The performance is shown for reaching to targets at the beginning, in the middle and at the end of learning for 6 trials from left to right. The black and red trails are for reaching without and with cerebellum, respectively. The blue color refers to the desired path.}
\label{fig:star_servo_}
\end{figure}

\begin{figure}
\centering
\begin{subfigure}{0.55\columnwidth}
\centering
\includegraphics[width=1.0\linewidth]{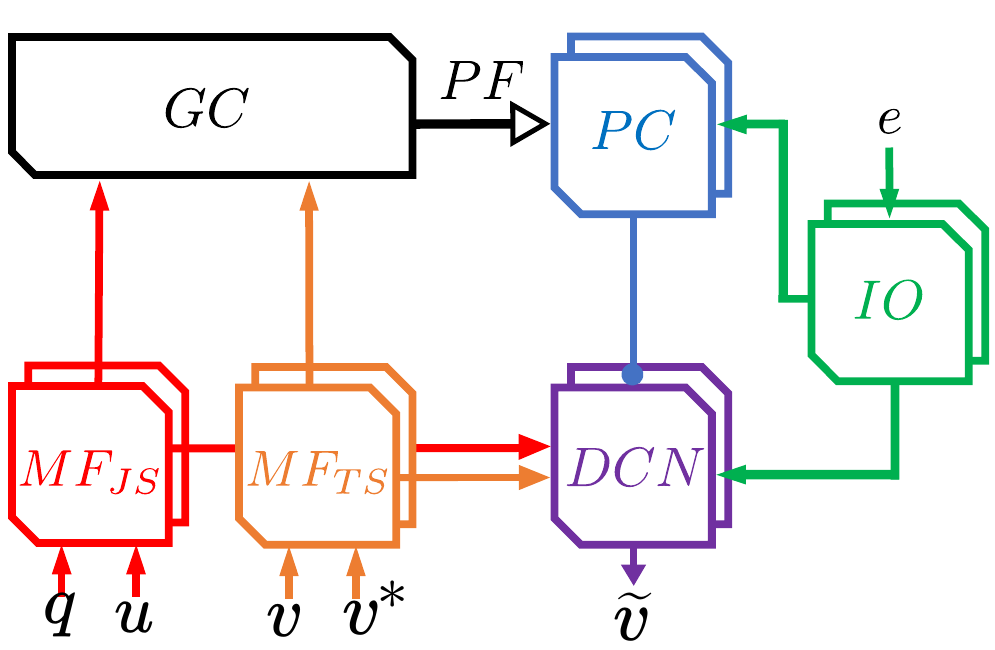} 
\caption{}
\label{fig:icra_net}
\end{subfigure}%
\begin{subfigure}{0.4\columnwidth}
\centering
\includegraphics[width=1.0\linewidth]{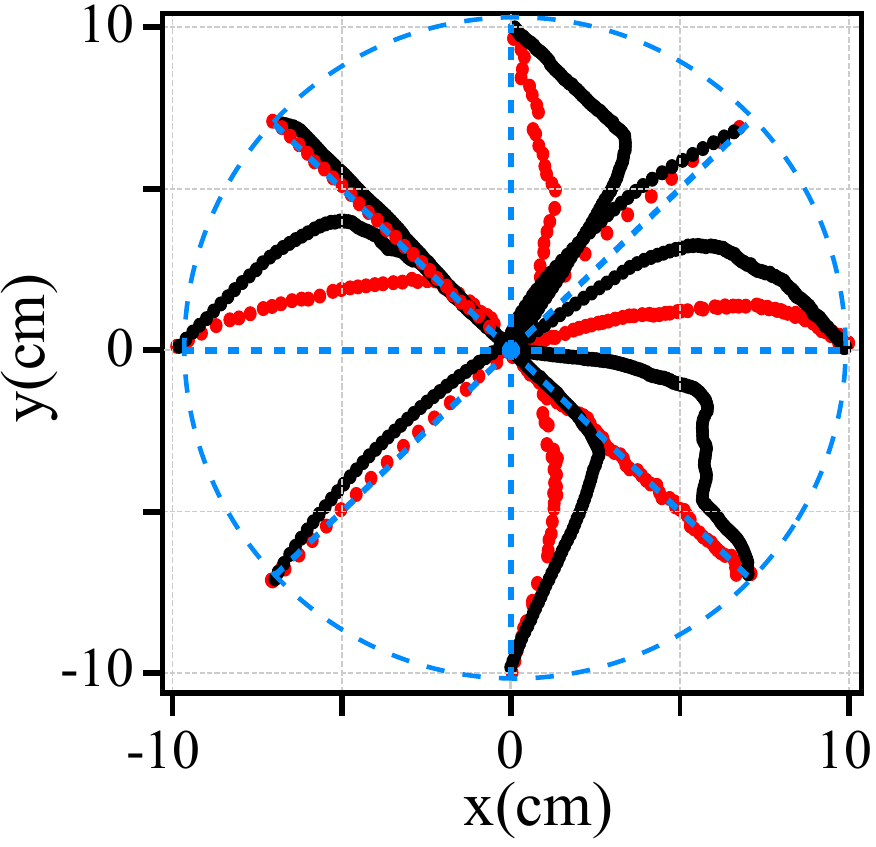}
\caption{}
\label{fig:star_servo_icra}
\end{subfigure}
\caption{(a)The network proposed in \cite{zahra2020vision} and (b) the results of radial reaching based after training the network for 8 repetitions.}
\label{fig:servo}
\end{figure}

 \noindent  starting from the same central point with the aim of testing the smoothness of movements of these patients and joint motion coordination compared to healthy persons. Similarly, the robot in this study is commanded to reach radially towards target points equally distributed in an eight-angled star shape with a 45$\degree$ internal angle and distant from the center 7 cm each. It can be concluded from the robot movements in the 8 directions while relying only on $DM$ in the left panel in Fig. \ref{fig:star_servo} that always the biggest error is presented in the first generated motor commands while the robot moves from rest. Thus, a small distance is defined for reaching to demonstrate the subsequent effects of the from-rest estimation error. The cerebellar network training is done by repeating the reaching motion to each of the 8 targets only six times with an improvement in the reaching motion as shown in the right panel in Fig. \ref{fig:star_servo} compared to what obtained after only 3 repetitions as shown in the middle panel. The maximum deviation recorded for repeating the reaching movement (after training ends) 10 times in each direction is reduced by a mean value of ca. 310\%, while the time for executing reaching movements is reduced by a mean value of ca. 235\%. The training and testing is done one by one for each direction. This outperforms the network developed in \cite{zahra2020vision}, as shown in Fig. \ref{fig:star_servo_icra} with a reduction of 55\% in the maximum deviation and 120\% in execution time, after training for 8 repetitions.

\subsection{Deformable Object Manipulation}\label{sec:obj_deform}
To demonstrate the ability of the developed cerebellar controller to facilitate learning different skills, an experiment is set for the robot to manipulate a deformable object. The contour of the deformable object is observed based on the color, and the moments are calculated such that:
\begin{equation}
    \mathcal{M}_{ij} = \sum_{x_1}\sum_{x_2} {x_1}^{(i)}{x_2}^{(j)}\varphi(x_1,x_2)
\end{equation}
where $\varphi$ gives the intensity of pixels.
The centroid $\mathcal{C}_x = (\mathcal{C}_x,\mathcal{C}_y)$ is then calculated in pixels based on the moment such that $\mathcal{C}_x=\mathcal{M}_{10}/\mathcal{M}_{00}$ and $\mathcal{C}_y=\mathcal{M}_{01}/\mathcal{M}_{00}$ as explained in \cite{hu1962visual}. 

The centroid of the object is then converted to the world coordinates using the intrinsic parameters of the camera \cite{Sturm2014}:
\begin{equation}
\label{eqn:eqlabel}
     x_1 = (\mathcal{C}_{x}-\mathcal{P}_{x})\frac{x_3}{\mathcal{F}}, \qquad
     x_2 = (\mathcal{C}_{y}-\mathcal{P}_{y})\frac{x_3}{\mathcal{F}}
\end{equation}
where $\mathcal{C}_{x}$ and $\mathcal{C}_{y}$ denote to the components of the centroid position in pixels, $x_3$ is the object's depth away from the camera, $\mathcal{P}_x$ and $\mathcal{P}_y$ denote the principal point and $\mathcal{F}$ is the focal length.

\begin{figure}
\centering
\includegraphics[width=\columnwidth]{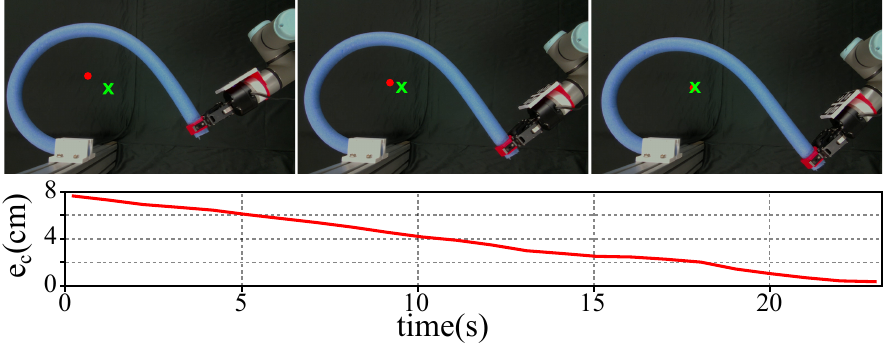}
\caption{The upper panel shows the shifting of the deformable object centroid (red dot) towards the target point (green cross). The lower panel demonstrates the almost linear decrease in the norm of the distance between the centroid and the target.}
\label{fig:servo_de}
\end{figure}

Similar to the motor babbling in case of end-effector tracking, in this case motor babbling is carried out and the corresponding value of centroid is recorded for the joint values to be used for training $DM$. It shall be noted that using the same parameters as those used in the previous experiment fails to develop a proper map to guide the robot. Thus, $DM$ alone fails to drive the robot to manipulate the object properly. However, the cerebellar model and the control architecture allow to drive the robot and develop the plastic synapses properly to guide the robot motion to deform the object properly as shown in Fig. \ref{fig:servo_de}. The robot is given 10 random targets in the studied work-space which are at least 5cm apart \\ 

\noindent from each other, only 3 of which are reached using the $DM$ with an average final error of around 7mm, while the 10 targets are reached successfully with an average final error of less than 4mm is achieved. While this study considers using the centroid as one feature to manipulate the deformable object, it may not be the optimal choice to achieve such task. Other features can be explored for future studies to give better candidates to reduce the high dimensional data needed to characterize a deformable object \cite{navarro2014visual}.

\section{Discussion and Conclusions}\label{sec:conclusions}
In this study, a biomimetic control system is developed based on the detailed microcircuit of the cerebellum. A spiking cellular-level forward cerebellar model is integrated with a differential map to helps improve the motion in terms of speed and deviation from the desired path. The learning is supervised by a teaching signal based on task-based sensory feedback, in contrast to most studies in the literature relying on joint-based errors as mentioned earlier. The forward model, acting as a Smith Predictor, then compares the expected output with the actual one to build an anticipation of error for the next cycle and provide corrections to the sensory feedback to the motor-cortex-like differential mapping network, where the spatial motion plan  is converted into motor commands. Both the angular and spatial accelerations are not included in this study as the robot moves at moderate speed with a light structure, hence the motion dynamics is not an effective factor. 
The network parameters are optimized using an ATPE based Bayesian Optimization. The optimization is carried out step by step to avoid the complexity of handling all the parameters at the same time. This allows to achieve the desirable performance while still maintaining the biological properties of the network components. This allows for future studies to use the detailed computational model to study cases with cerebellar damage in which monitoring the activity of all the neurons simultaneously is not feasible. 

The obtained results show that cerebellum acts to reduce the deviation from the target path and the execution time. Additionally, it demonstrates the ability to learn new skills such as deformable object manipulation based on the error in task performance. The developed model demonstrates the ability to reduce the error in a certain direction in only few repetitions indicating the fast convergence of learning, and the suitability to be further developed for real-time adaptive robot control in multiple scenarios and applications. Moreover, it shall be noted from the conducted experiments that the estimation error in $DM$ is not uniform across the map. This is related to the way of collecting and introducing data for training. This is analogous to having areas mapping different body parts in the motor cortex having different density of neurons depending on how frequent and accurate are the motions generated by each body part.  Hence, providing sensory corrections from $CB$ allows $DM$ to improve the quality of the motor output and points to the possibility of transfer of learning by having the cerebellum correcting the motion, and thus improving the quality of the training data introduced to the motor cortex.
The radial reaching experiment obtained with the robot arm displays a fair similarity with the observations in \cite{Fortier2002CEREBELLARAD}.
This study considers only the planar motion at the end-effector as those carried out to test the motion of patients suffering from cerebellar damage. Future studies shall include non-planar motions and incorporate dynamics/forces acting at the end-effector \cite{capolei2019biomimetic}.
Moreover, future models would include more features, where developing highly detailed models would allow identifying cerebellar dysfunctions and studying lesions at the cellular-level, which may not be possible using current state-of-art techniques.

\section{Funding}
This research work was supported in part by the Research Grants Council (RGC) of Hong Kong under grant number 14203917, in part by PROCORE-France/Hong Kong Joint Research Scheme sponsored by the RGC and the Consulate General of France in Hong Kong under grant F-PolyU503/18, in part by the
Key-Area Research and Development Program of Guangdong Province 2020 under project 76 and in part by The Hong Kong Polytechnic University under grant G-YBYT.


\bibliography{biblio.bib}
\bibliographystyle{IEEEtran}

\end{document}